\documentclass[a4paper, 11pt]{article}
\usepackage{natbib}
\usepackage[ruled]{algorithm2e}
\usepackage[T1]{fontenc}
\usepackage[latin1]{inputenc}
\usepackage{amsmath, amssymb}
\usepackage{graphicx}
\usepackage{booktabs}
\usepackage{geometry}
\usepackage{hyperref}

\hypersetup{pdfauthor={Tobias Buer, J\"org Homberger, Hermann Gehring},     
    pdfsubject={A collaborative ant colony metaheuristic for distributed multi-level lot-sizing},   
    pdfcreator={Tobias Buer},  
    pdfproducer={Tobias Buer}, 
    pdfkeywords={Group decisions and negotiations; metaheuristics; collaborative planning; ant colony optimization; lot sizing}, 
    colorlinks=true, 
    linkcolor=blue,
    citecolor=blue,
    urlcolor=blue, 
    bookmarksopen=true, 
    bookmarksopenlevel=2, 
    bookmarksnumbered=true, 
}
\begin{document}

\title{A collaborative ant colony metaheuristic for \\ distributed multi-level lot-sizing}

\author{Tobias Buer$^{a}$\thanks{Corresponding author. Email: tobias.buer@uni-bremen.de
\vspace{6pt}} \and J\"org Homberger$^{b}$ \and Hermann Gehring$^{c}$}
\date{
\vspace{6pt}  $^{a}${\em{Chair of Logistics, University of Bremen, 28334~Bremen, Germany}} \\
$^{b}${\em{Faculty of Geomatics, Computer Science and Mathematics, Stuttgart University of Applied Sciences, 70174~Stuttgart, Germany}} \\
$^{c}${\em{Department of Information Systems, University of Hagen, 58084~Hagen, Germany}}\\\vspace{6pt}}

\maketitle

\begin{abstract}
\noindent
The paper presents an ant colony optimization metaheuristic for collaborative planning. Collaborative planning is used to coordinate individual plans of self-interested decision makers with private information in order to increase the overall benefit of the coalition.
The method consists of a new search graph based on encoded solutions.
Distributed and private information is integrated via voting mechanisms and via a simple but effective collaborative local search procedure.
The approach is applied to a distributed variant of the multi-level lot-sizing problem and evaluated by means of 352 benchmark instances from the literature.
The proposed approach clearly outperforms existing approaches on the sets of medium and large sized instances.
While the best method in the literature so far achieves an average deviation from the best known non-distributed solutions of 46 percent for the set of the largest instances, for example, the presented approach reduces the average deviation to only 5 percent. \bigskip

\bigskip
\end{abstract}

\section{Introduction and literature review}
A coalition of decision makers with private information that requires collaborative production planning is considered.
Each of the decision makers, hereafter referred to as agents, is selfish and seeks to implement his or her local optimal production plan to minimize his or her local costs.
However, the agents could improve their situation by coordinating their individual production plans.
The cost savings due to a better global production plan could be allocated to the agents in order to overcompensate them for deviating from their local optimal production plans.
From a more general point of view, the problem of finding an appropriate allocation of cooperative gains is dealt with by cooperative game theory. The special situation of joint planning of selfish agents in case of private information considered here is the subject of collaborative operations planning \citep{Dudek_2005}.

The state of the art in collaborative planning is discussed by \citet{Frayret_2009} and by \citet{Stadtler_2009}. Overviews of automated negotiation approaches relevant to collaborative planning are given by \citet{Lomuscio_2003} and \citet{Stroebel_2003}. Applications of collaborative planning to supply chains are discussed by \citet{Ertogral_2000}, and \citet{Fink_2004a, Fink_2006}, for example.

The proposed collaborative planning approach is applied to a distributed variant of the well-known multi level uncapacitated lot sizing problem (MLULSP) introduced by \citet{Yelle_1979}. The \emph{distributed} MLULSP (DMLUSLSP) was presented by \citet{Homberger_2010} and assumes several local and selfish agents with private information instead of a single agent with full information. The DMLULSP covers some important features of real world problems.
There are several final products, a multi level production structure, and a trade-off between inventory and setup costs. Coordination is difficult, because there are agents with private information and conflicting objectives.
Finally, the problem is computationally challenging, as it is NP-hard for general product structures \citep{Arkin_1989}.
As to computational experiments, a variety of benchmark instances is available in the literature which allow for comparative test including other approaches.

Lot Sizing Problems are of high relevance in modern supply chains \citep{Voss_2006}.
Surveys of the rich literature on lot sizing problems are given for example by \citet{Drexl_1997}, \citet{Jans_2007}, or \cite{Winands_2011}.
Integration of lot sizing with cooperative game theory is discussed by \citet{Drechsel_2010, Drechsel_2011}.
Metaheuristic solution approaches for the non-distributed MLULSP are amongst others presented by \citet{Dellaert_2000, Dellaert_2003}, \citet{Pitakaso_2007}, \citet{Homberger_2008}, \cite{Xiao_2011}, and \cite{Xiao_2012}.

The literature discusses at least two \emph{distributed} lot sizing variants. The one studied by \citet{Lee_2007, Lee_2007b} is based on auctions and has been tested on small size instances. For the other, the distributed MLULSP, several metaheuristic approaches that integrate concepts from group decision making are known in the literature: a simulated annealing approach \citep[SA, ][]{Homberger_2010}, an approach based on an evolutionary strategy \citep[ES, ][]{Homberger_2011} and one based on ant colony optimization \citep[ACO, ][]{Homberger_2010a, Homberger_2011a}. These approaches differ by the used metaheuristic principle, the applied negotiation or voting mechanisms, and the relevant objective functions (minimizing total global cost or maximizing fairness). ES is especially useful for maximizing fairness. SA and ACO strive to minimize total global cost. In terms of solution quality, ACO outperforms SA on medium sized instances (40 and 50 items with 12 and 24 periods) but not on small or large instances (500 items, and 36 or 52 periods). Furthermore, ACO is significantly slower than SA on small, medium, and large instances.

The aim of this paper is to present a new collaborative ant colony metaheuristic (CACM) for solving the DMLULSP under global total cost minimization. Compared to \citet{Homberger_2011a} and the other mentioned approaches, the proposed CACM incorporates the following new conceptual and methodical ideas:
\begin{enumerate}
  \item Encoded solutions are represented by a new and simplified search graph.
  Let $m$ denote the number of items to produce and let $n$ denote the number of periods.
  The search graph proposed by \citet{Homberger_2009a} and used in \citet{Homberger_2011a} requires $m(n+1)$ nodes and $\frac{1}{2}mn(n + 1)$ directed edges while the new search graph uses $2mn+1$ nodes but only $4mn-2$ edges. The new search graph facilitates learning of pheromone values and therefore makes it easier to discover promising paths which lead to high quality solutions.

  \item Individual preferences of the agents -- which are private -- are aggregated by a voting mechanism and by a new collaborative local search procedure.
       Usually, to be able to reach a (near) optimal solution fast, an ant colony metaheuristic requires heuristic information on the problem, for example, the costs of the agents. However, this operational information is sensitive and therefore private. The new approach exploits this information indirectly, which enables to guide the search via heuristic information and, nevertheless, keeps operational information private.

  \item The new approach does without rank based voting mechanisms. This reduces computational effort and increases robustness, that is, there are less opportunities for undesired strategic behavior of the agents.

\end{enumerate}
As it is shown below, the incorporation of these concepts has a high positive impact on the solution quality of the resulting approach.

The rest of the paper is organized as follows.
In Section~\ref{sec:problem} the considered distributed multi-level lot sizing problem (DMLULSP) is characterized. Section~\ref{sec:aco} describes the developed collaborative ant colony metaheuristic for solving the DMLULSP. Subject to Section~\ref{sec:aco} is the evaluation of the new approach by means of a computational study with 352 benchmark instances. Finally, concluding remarks are given in Section~\ref{sec:conclusion}.

\section{A distributed multi-level uncapacitated lot-sizing problem}
\label{sec:problem}

\subsection{Classical centralized problem formulation}
The distributed lot-sizing model studied in this paper extends the well-known multi-level uncapacitated lot sizing problem (MLULSP, cf. \citealt{Yelle_1979}, \citealt{Steinberg_1980}, \citealt{Dellaert_2000}).
Therefore, the MLULSP is presented first using the notation given in Table~\ref{tab:notation}. The MLULSP assumes a single decision maker who is aware of all information required for planning, especially, the setup costs and inventory holding costs per item and period.
A formal description of the MLULSP is given by the formulas (\ref{eq:mlulsp_objective}) to (\ref{eq:setup}).

\begin{table}
\centering
\caption{Notation for the DMLULSP.}
\label{tab:notation}
\vspace{5pt}
\small
\begin{tabular}{lp{11.5cm}} \toprule
\multicolumn{2}{l}{\textbf{Problem parameters}} \\ \midrule
$M$ & a sufficiently large number \\

$m$ & number of items \\

$n$ & number of production periods \\

$I$ & set of items, $I = \{1, \ldots, m\}$ \\

$T$ & set of possible production periods, $T = \{1, \ldots, n\}$ \\

$s_i$ & setup costs per period for item $i \in I$ \\

$h_i$ & inventory holding costs per period and per unit of item $i \in I$ \\

$t_i$ & lead time required to assemble, manufacture, or purchase item $i \in I$ \\

$r_{ij}$ & number of items $i$ required to produce one unit of item $j$ with $i,j \in I,\, i \neq j$ \\

$\Gamma^{+}(i) \subset I$  & set of all direct successors of item $i \in I$ \\

$\Gamma^{-}(i)  \subset I$ & set of all direct predecessors of item $i \in I$ \\

$d_{it}$ & exogenous demand (unit of quantity) for item $i \in \{j \in I | \Gamma^{+}(j) = \emptyset\}$ in period $t \in T$ \\

 \midrule

\multicolumn{2}{l}{\textbf{Decision variables}} \\ \midrule

$d_{it}$ & endogenous demand (unit of quantity) of item $i \in \{j \in I | \Gamma^{+}(j) \neq \emptyset\}$ in period $t \in T$  \\

$l_{it}$ & inventory (unit of quantity) of item $i \in I$ at the end of period $t \in T$ \\

$x_{it}$ & lot size (unit of quantity) of item $i \in I$ in period $t \in T$ \\

$y_{it}$ & binary setup decision, $y_{it} = 1$ if item $i \in I$ is produced in period $t \in T$ and $y_{it} = 0$ otherwise \\
\bottomrule
\end{tabular}

\end{table}

\begin{align}
\label{eq:mlulsp_objective}
\quad \quad \min f^{nd}(y)  \quad & =  \sum_{i \in I} \sum_{t \in T}{(s_i \cdot y_{it}  +  h_i \cdot l_{it})} & 	\\
\label{eq:balance}
\text{s. t.} \quad \quad 	l_{it} & = l_{i,t-1} +  x_{it} - d_{it}\,, 	 &	\forall i \in I,\, \forall t \in T, \\
\label{eq:invfp}
		l_{i,0} & =  0\,, &	\forall i \in I,	\\
\label{eq:invnn}
		l_{it} & \geq  0\,, &	\forall i \in I,\, \forall t \in T \setminus \{0\},	\\
 \label{eq:demand}
		d_{it} & = \sum_{j \in \Gamma^{+}(i)}{r_{ij} \cdot x_{j,t+t_i}}\,, 				&	\forall i \in \{j \in I \mid \Gamma^{+}(j) \neq \emptyset \},\, t \in T,	 \\
\label{eq:bigM}
		x_{it} - M \cdot y_{it} & \leq 0\,, &	 \forall i \in I,\, \forall t \in T,	\\
\label{eq:lotsizenn}
		x_{it} & \geq 0\,, &	\forall i \in I,\, \forall t \in T,	\\
\label{eq:setup}
		y_{it} & \in \{0, 1\}\,, & \forall i \in I,\, \forall t \in T.
\end{align}

The goal of the MLULSP is to minimize the total costs $f^{nd}$ of a single, central decision maker. These are expressed by the objective function (\ref{eq:mlulsp_objective}) which sums up the setup costs and the stockholding costs for all items $i \in I$ over all periods $t \in T$.
The inventory balance is guaranteed by (\ref{eq:balance}).
For all items, the inventory of the first period $t=0$ is zero (\ref{eq:invfp}) and for remaining periods non-negative (\ref{eq:invnn}).
For each period, the demands $d_{it}$ for the level zero items $i \in I$ with $\Gamma^{+}(i)=\emptyset$ are given.
The demands for the remaining items are determined by (\ref{eq:demand}).
These constraints ensure that the production of item $j$ in period $t+t_i$ triggers a corresponding demand $d_{it}$ for all $i \in \Gamma^{-}(j)$, that is a demand for each item $i$ that is preceding item $j$ in the bill of materials.
Without loss of generality, $r_{i,j}=1$ is assumed in (\ref{eq:demand}). 
The lot size $x_{it}$ is non-negative (\ref{eq:lotsizenn}).
If $x_{it} > 0$, that is, item $i$ is produced in period $t$, then $y_{it} = 1$, otherwise $y_{it} = 0$.
This is enforced by the constraints (\ref{eq:bigM}) and (\ref{eq:setup}).

\begin{figure}
 \centering
 \includegraphics[scale=0.8]{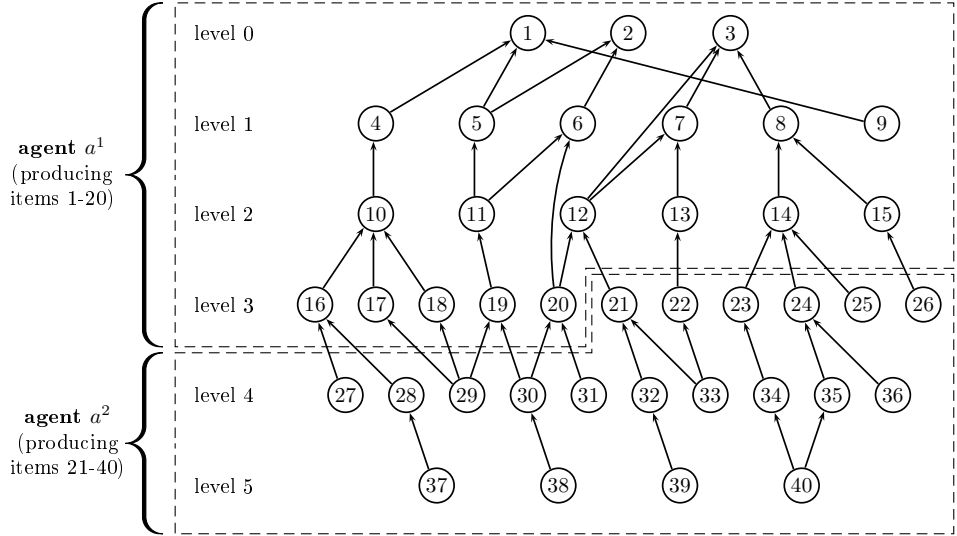}\\
 \caption{Example of a product structure and a partition of the items to two agents \citep[cf.][]{Homberger_2010}.} \label{fig:itemstructure}
\end{figure}

\cite{Arkin_1989} have shown that the MLULSP is NP-hard for general multi-level product structures, i.\,e., for product structures, where each item can have multiple successors and predecessors like in Figure~\ref{fig:itemstructure} \citep{Salomon_1993}.
Figure~\ref{fig:itemstructure} shows an example for a graphical representation of a product structure with $m = 40$ which follows the literature \citep{Bookbinder_1990}. The production dependencies between items are indicated by arrows. Furthermore, items are classified by decreasing production level from top to bottom. Each end product $i \in I$ with $\Gamma^{+}(i)=\emptyset$ is located at the highest level which is level 0. Each common part is located at the lowest level at which it is used anywhere in the product structure \citep{Dellaert_2000}. The allocation of items to a group of agents also shown in Figure~\ref{fig:itemstructure} is described next.

\subsection{Extension to a distributed group decision problem formulation}

Following \citet{Homberger_2010}, the MLULSP is now extended to a distributed group decision problem which is called distributed multi-level uncapacitated lot-sizing problem  (DMLUSLSP). Instead of a single decision maker or agent who is responsible to produce all items $I$, the responsibility is jointly assigned to a group of agents $A$. Each agent $a \in A$ is responsible to produce the set of items $I_a$ with $\bigcup_{a \in A} I_a = I$ and $\bigcap_{a \in A} I_a = \emptyset$. For instance, an agent might represent a profit center or an independent company and the agents might interact in a supply chain.

All agents are autonomous and self-interested. Therefore, the individual objective function $f_a$ of agent $a \in A$  is to minimize his or her total local costs for producing the items $I_a$, that is,

\begin{align}
\min f_a(y) = \sum_{i \in I_a} \sum_{t \in T} (s_i \cdot y_{it}  +  h_i \cdot l_{it}). \label{eq:local_evaluation}
\end{align}

Furthermore, for all agents $a \in A$ and for all items $i \in I_a$, the DMLULSP assumes \emph{asymmetric information} regarding the cost parameters $s_i$ and $h_i$. These cost parameters are private information. That is, at the initiation of planning they are only known to agent $a$ and during planning agent $a$ does not want to reveal these to other agents. If this information would be common knowledge, price negotiations between agents might be negatively affected.
On the other hand, \emph{symmetric information} is assumed regarding the bill of materials. This assumption can be justified by some kind of common industry knowledge or joint development of products (e.g. collaborative engineering).

The DMLUSLSP consists of the constraints (\ref{eq:balance}) to (\ref{eq:setup}) and the objective function (\ref{eq:global_evaluation}) which minimizes the total global costs

\begin{align}
\min f(y) = \sum_{a \in A} f_a. \label{eq:global_evaluation}
\end{align}

In comparable scenarios, the total global costs are also considered of interest by \citet{Fink_2004a}, \citet{Zimmer_2004}, \citet{Jung_2005}, and \citet{Dudek_2005, Dudek_2007}.
In the following, the function (\ref{eq:global_evaluation}) is simply referred to as total cost function or central cost function, the function (\ref{eq:local_evaluation}) is also denoted as individual or local cost function of agent $a \in A$. Usually, the minimization of the individual cost functions conflicts with the objective of minimizing the total costs.
In order to resolve these conflicts and support the agents in agreeing on a joint production plan in order to produce all items $I$, the solution approach proposed in the following section uses collaborative local search and different voting mechanisms.

\section{A collaborative planning metaheuristic based on ant colony optimization} \label{sec:aco}

\subsection{Overview}
A metaheuristic solution approach based on ant colony optimization (ACO) is presented that especially takes into account the distributed nature of the DMLULSP.  The distributed solution approach is denoted as collaborative ant colony metaheuristic (CACM); an overview is given by Alg.~\ref{alg:caco}.
The metaheuristic CACM works on an \emph{encoded} representation of a DMLULSP solution which is described in Sect.~\ref{sec:representation}.
Components of an encoded solution are represented by a search graph $G$ which is traversed by artificial ants in order to construct encoded solutions.
Sect.~\ref{sec:graph} introduces the search graph $G$ and Sect.~\ref{sec:construct} presents the stochastic ant construction procedure.
Each constructed solution is improved by a collaborative local search heuristic (cf. Sect.~\ref{sec:cls}).
Both, the used graph structure and the local search heuristic incorporate new concepts compared to the ant metaheuristic of \citet{Homberger_2011a}.
After a set $B$ of encoded solutions has been constructed, the agents negotiate about the acceptance of a solution in $B$ by casting votes according to one of several voting rules described in Sect.~\ref{sec:voting}.
Finally, the pheromone information related to the arcs of the search graph are updated according to the present accepted solution $e^*$  (Sect.~\ref{sec:pherup}). That is, in the next iteration of CACM ants use a different probability distribution in order to construct a new encoded solution.

\begin{algorithm}
\caption{Collaborative Ant Colony Metaheuristic} \label{alg:caco}
\DontPrintSemicolon

\SetKw{Mod}{mod}

\SetKwFunction{Construct}{constructEncodedAntSolution}
\SetKwFunction{Update}{updatePheromoneValues}
\SetKwFunction{Vote}{voting}
\SetKwFunction{CLS}{collaborativeLocalSearch}

\KwIn{no. of items $m$, no. of periods $n$, set $A$ of agents, no. of generated solutions  $\overline{s}$, ballot size $b$, min and max pheromone values $\tau^{min}$ and $\tau^{max}$, evaporation rate $\rho$, intensification rate $\sigma$}
\KwOut{jointly accepted encoded solution $e^*$}
\BlankLine
initialize search graph $G$ and pheromone values $\tau$
\tcp*[r]{Section~\ref{sec:graph}}
$e^{*}_{it} \leftarrow 0, \quad i = 1, \ldots, m; \; t = 1, \ldots, n$ \;
\For{$s \leftarrow 1 $ \KwTo $ \overline{s}$}{
    $B \leftarrow \{\}$ \;

    \For{$b \leftarrow 1 $ \KwTo $ \overline{b}$}{
        $e \leftarrow \Construct(G, \tau ) $
        \tcp*[r]{Section~\ref{sec:construct}}

        $e \leftarrow \CLS(e, A) $ \tcp*[r]{distributed, Section~\ref{sec:cls}}

        $B \leftarrow B \cup \{e\}$ \;
        $s \leftarrow s + 1 + m (n-1)$
    }
    $ e^* \leftarrow \Vote(B, A) $
    \tcp*[r]{distributed, Section~\ref{sec:voting}}

    $ \tau \leftarrow \Update(e^*, \tau, \tau^{min}, \tau^{max}, \rho, \sigma)$
    \tcp*[r]{Section~\ref{sec:pherup}}
}
\Return $e^*$ \;
\end{algorithm}

CACM requires several input parameters. The problem data include the number of items $m \in \mathbb{N}$, the number of periods $n \in \mathbb{N}$, and the set $A$ of agents. The termination of CACM is controlled by the number of generated solutions $\overline{s} \in \mathbb{N}$. The ballot size $b \in \mathbb{N}, b \Leftrightarrow |B|,$ specifies the number of solutions which can be negotiated by the agents $A$ in a single election. The remaining parameters influence the update of the pheromone information on the arcs of the search graph $G$. The minimum and maximum pheromone values are defined by $\tau^{min}, \tau^{max} \in \mathbb{N}$. The pheromone evaporation rate and the pheromone intensification are given by $\rho$ and $\sigma$ with $0 \leq \rho,\sigma \leq 1$.

The solution approach CACM can deal with asymmetric information distributed among the agents. CACM is executed by a neutral mediator.
In order to construct encoded solutions, the mediator has to be aware of the items $I$ to produce and of the possible production periods $T$.
The agents do not have to reveal private information like costs or free production capacities, neither to the mediator nor to other agents. To control the search process in order to find a jointly accepted solution $e^*$, the mediator has to interact with the agents in both steps marked as \emph{distributed} in Alg.~\ref{alg:caco}.
CACM outputs a jointly accepted solution $e^*$.

\subsection{Representation of a solution} \label{sec:representation}

\subsubsection{Solution encoding} \label{sec:encoding}

For the MLULSP, the decision variables $x_{it}$ represent the lot size of item $i$ produced in period $t$. The binary decision variables $y_{it}$ represent the setup decisions which indicate wether item $i$ is produced in period $t$ at all. Because of the structure of the MLULSP, it is possible to determine the optimal lot sizes $x_{it}$, if the optimal setup decisions $y_{it}$ are known \citep{Dellaert_2000}.

Therefore, the proposed CACM focuses its search effort to approximate the set of optimal setup decisions. However, CACM does not operate on the $y_{it}$ directly, but indirectly on an encoded representation of the setup decisions. For each binary setup decision variable $y_{it}$, there is an encoded binary setup decision variable $e_{it}$ following the suggestion of \citet{Homberger_2008}. Both variables are related as follows:

\begin{align}
(e_{it}=0) &\Leftrightarrow (y_{it} = 0), \\
(e_{it}=1) &\Rightarrow (y_{it} = 0 \vee y_{it} = 1).
\end{align}

Accordingly, the variables $e_{it}=1$ indicate a \emph{possible}, but not mandatory, production of item $i \in I$ in period $t \in T$. Another ACO approach that also uses encoded decision variables is suggested by \citet{Fink_2007}, for example. Whether a possible production takes place if $e_{it}=1$, can be deduced by the following decoding rule.

\subsubsection{Solution decoding} \label{sec:decoding}

An encoded solution $e \in \{0,1\}^{m \times n}$ is transformed to a DMLULSP solution in a two steps procedure.
The product level of an item $i$ determines the order in which demand $d_{it}$, lot size $x_{it}$, and inventory $l_{it}$ are calculated. The procedures starts with the lowest level items (final products) and advances with items of increasing product level.
That is, the demands for the final products (level 0) are determined first, the demands for level 1 items are calculated next, and so on (cf. \citet{Homberger_2008})

Hence, the items are considered in the sequence of their non-decreasing product level.
If item $i$ is a final product (level 1), then $d_{it}$ is given for all $1 \leq j \leq T$. For each non-final product $i$ the demand can be calculated according to equation (\ref{eq:demand}) using to the demands of all successors of $i$ indicated by $\Gamma(i)$. All $d_{kj}, k \in \Gamma(i)$ have already been calculated, because of the used calculation sequence.
By means of the second decoding step, the setup decisions, lot sizes, and stored quantities are calculated.
One should note, that with the used encoding and decoding rules, each encoded solution with a $e_{i1}=1$ for $i = 1,\ldots, N$ represents a feasible decoded solution in any case.

\subsection{Definition of a search graph} \label{sec:graph}
\subsubsection{Search graph}

\begin{figure}
\vspace{36pt}
\centering
\includegraphics[scale=1]{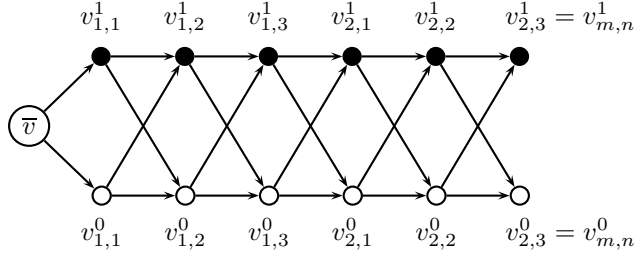}
\caption{Example of a search graph for two items and three periods.}
\label{fig:searchgraph}
\end{figure}

The ACO principle is guided by the image of an ant that constructs a feasible solution by traversing a graph which consists of solution components specific to the problem.
A solution is represented by the path chosen by the ant.
In order to enable such an artificial ant to construct an \emph{encoded} solution for the DMLULSP, a directed graph $G = (V,E)$ with a set $V$ of nodes and a set $E$ of directed edges is used (see Figure~\ref{fig:searchgraph}).
The node set $V$ consists of three disjunct subsets, i.\,e., $V = \{\overline{v}\} \cup V^1 \cup V^0$.
The starting point of an ant is the initial node $\overline{v}$.
For each of the two possible values of an encoded decision variable $e_{ij}$ there exists a node.
The encoded decision to produce item $i$ in period $j$ ($e_{ij}=1$) is represented by node $v^1_{ij} \in V^1$.
The encoded decision $e_{ij}=0$ is represented by node $v^0_{ij} \in V^0$. Therefore, $V^1 \cap V^0 = \emptyset$ and $|V^1| = |V^0|$.
The nodes in $V^1$ are called black nodes, the nodes in $V^0$ are called white nodes.

The set of directed edges $E$ is defined as follows.
The initial node $\overline{v}$ is associated with two directed edges $(\overline{v}, v^1_{1,1})$ and $(\overline{v}, v^0_{1,1})$.
Furthermore, each node $v \in V \setminus \{v^1_{m,n}, v^0_{m,n}\}$ acts as the origin of exactly two directed edges.
From a black node two edges originate, one points to another black node and one points to a white node; analogously, from a white node one edge points to a black node and one edge points to another white node.
More precisely, let $a, b \in \{0,1\}, a \neq b$ and $v^a \in V^a, v^b \in V^b$. The pair of edges originating from $v^a_{ij}$ is defined as $(v^a_{ij}, v^a_{i,j+1})$ and $(v^a_{ij}, v^b_{i,j+1})$, provided that $j < n$. Else, if $j = n$, then the pair of edges is defined as $(v^a_{ij}, v^a_{i+1,1})$ and $(v^a_{ij}, v^b_{i+1,1})$.

To ease presentation, two subsets of directed edges $E^1$ and $E^0$ are introduced. $E^1$ contains those edges that point to a black node and $E^0$ contains those edges that point to a white node.

\begin{align} \label{eq:pherIni}
E & = E^1 \cup E^0 \text{ with } E^1 \cap E^0 = \emptyset, \\
E^1 & := \{(i,j) \in E | i \in V \wedge j \in V^1\},  \\
E^0 & := \{(i,j) \in E | i \in V \wedge j \in V^0\}.
\end{align}

An \emph{encoded} solution to the DMLULSP is represented by all nodes $v \in V^1 \cup V^0$ on any path through $G$ that starts at node $\overline{v}$ and ends either at node $v^1_{m,n}$ or at node $v^0_{m,n}$.

\subsubsection{Initialization of pheromone trail}

The sequence of directed edges traversed by an ant depends on the pheromone value $\tau_{(i,j)}$ placed on each edge $(i,j) \in E$. Higher pheromone values $\tau_{(i,j)}$ on an edge $(i,j)$ increase the probability that an ant traverses edge $(i,j)$.  \citet{Homberger_2009a} proposed an ACO approach for a (non-distributed) MLULSP on a different search graph and successfully pursued the idea to initially construct each possible solution with equal probability ('balanced path'). In contrast to that idea, the goal of this approach is to initially construct unbalanced paths that predominantly consist of black nodes. Following the max-min ant system proposed by \citep{Stuetzle_2000}, a minimum and a maximum pheromone value for each edge $(i,j) \in E$ are defined as $\tau^{min}$ and $\tau^{max}$. The initial pheromone value $\tau_{(i,j)}$ for each edge $(i,j) \in E^0$ is set to $1$, and the initial pheromone value for $(i,j) \in E^1$ is set to $\tau^{max}-1$.

\subsection{Construction of a solution} \label{sec:construct}

\subsubsection{Component selection probability}
In order to construct an encoded solution $e$, the graph $G$ is \emph{traversed} by an ant along the directed edges.
The traversal of $G$ starts at the initial node $\overline{v}$.
Until the ant arrives at node $v^1_{m,n}$ or $v^0_{m,n}$, the ant decides randomly at each node $v \in V \setminus \{v^1_{N,T}, v^0_{N,T}\}$ which of the two possible edges $(v,i) \in E^1$ and $(v,j) \in E^0$ to choose.
The random decision depends on the pheromone intensity $\tau_{(i,j)}$ of an edge $(i,j) \in E$. The probability to choose edge $(i,j)$  is

\begin{align} \label{eq:prob}
  &p_{(i,j)} = \frac{\tau_{(i,j)}}{\tau^{max}} \text{ with } \tau_{(i,j)} + \tau_{(i,k)} = \tau^{max} \text{ and } (i,j) \in E^1 \wedge (i,k) \in E^0\,.
\end{align}

Hence, an according production for the first period of demand is planned and the calculation of a feasible solution is guaranteed \citep{Dellaert_2000}.

\subsubsection{Heuristic information}

Typically, ACO approaches from the literature additionally bias the selection probability of a component (here, an edge $(i,j) \in E$) via heuristic information. In the present case, however, the DMLULSP is a distributed problem with asymmetric information which is solved by a mediator. Because the mediator is neutral, he or she cannot take into account heuristic information that would depend on an individual agent and therefore privilege (or penalize) this agent. For this reason, equation (\ref{eq:prob}) does without local heuristic information.

\subsection{Collaborative local search heuristic} \label{sec:cls}

An encoded solution $e$ constructed by an ant is improved by a collaborative local search heuristic (CLS, cf. Alg.~\ref{alg:cls}). In this heuristic, a move is only executed, if each agent $a \in A$ agrees.
A move is defined as a bit shift, that is, for all $i \in I$ and all $t \in T, t > 1$, the value of $e_{it}$ is shifted from zero to one or one to zero, respectively. Each shift leads to new solution $e'$. The solution $e'$ is submitted to all agents $a \in A$. If and only if $e'$ has lower local costs $f_a(e')$ compared to $f_a(e)$ for each agent $a \in A$, the next move is executed on $e'$ and so on.

\begin{algorithm}
\caption{Collaborative local search} \label{alg:cls}
\DontPrintSemicolon
\SetKw{App}{veto}

\KwIn{encoded solution $e$, set of agents $A$}
\KwOut{possibly modified encoded solution $e$}
\BlankLine
\ForEach{$t \leftarrow 2, \ldots, n$}{

    \ForEach{$i \leftarrow 1, \ldots, m$}{
        $e' \leftarrow e $ \;
        $e'_{it} \leftarrow \neg e_{it}$ \tcp*[r]{bit shift}
        $s \leftarrow s + 1$ \;
        veto $\leftarrow false$ \;
        \ForEach(\tcp*[f]{complete approval voting}){$a \in A$}{
            \lIf{$f_a(e') > f_a(e)$}{ veto $ \leftarrow true $ }  \tcp*[r]{each agent $a$ can veto $e$}
        }
        \lIf{veto = false}{ $e \leftarrow e'$ } \;
    }
}
\Return $e$ \;
\end{algorithm}

In CACM (cf. Alg.~\ref{alg:caco}), the counter of generated solutions $s$ is increased by $1+m(n-1)$, as there are $m(n-1)$ possible moves and each move applied to a generated ant solutions leads to a new solution. Furthermore, the final solution $e$ generated by CLS is in CACM again submitted to all agents. This time, the agents compare $e$ to the jointly accepted solution $e^*$ and decide via voting, if $e$ should become the new jointly accepted solution $e^*$.

\subsection{Distributed decision making by means of voting} \label{sec:voting}

\subsubsection{Complete approval voting}
In approval voting each agent can vote for as many candidate solutions in $B$ as he or she wants and the solution with the most approvals wins \citep{Brams_1978}.
For the problem at hand, a modified and restricted version of approval voting is used which is denoted by the term complete approval voting. An encoded candidate solution $e$ is only accepted as winner, if \emph{all} agents $a \in A$ approve $e$.
Therefore, by accepting $e$, no agent increases his or her current local costs $C^*_a$, i.\,e., $C_a(e) \leq C_a^*, \; \forall a \in A$.
For complete approval voting, the ballot size is restricted to $b=1$.
Although this voting rule is quite simple, it has not been used in the previous approaches of \citet{Homberger_2010, Homberger_2011} and \citet{Homberger_2011a}.

\subsubsection{Adapted Borda maximin voting rule}

Each agent $a \in A$ ranks the candidate solutions $e \in B$ according to his or her preferences.
A ranking of $e \in B$ by agent $a$ is given by $e^a_{p} \succ e^a_{p-1} \succ \ldots \succ e^a_{1}$ where candidate solution $e^a_p$ is preferred most by agent $a$ and receives $p$ points. Solution $e^a_{p-1}$ is ranked second and receives $p-1$ points, candidate solution $e^a_1$ is preferred least and receives $1$ point.
Given all rankings by all agents $a \in A$, the Borda maximin rule selects that candidate solution $e'$ which maximizes the lowest number of points assigned by an agent to a candidate solution.

Finally, the solution $e'$ identified by Borda voting is compared to the jointly accepted solution $e^*$ by means of complete approval voting. The winning solution becomes the new jointly accepted solution $e^*$.

\subsubsection{Rawls or minimax voting rule}
The minimax voting rule, introduced by \citet{Rawls_1971}, minimizes the maximum local cost of an agent. In contrast to the voting rules Borda maximin and complete approval, the Rawls voting rule requires the agents to reveal their local cost to the mediator. For each encoded candidate solution $e \in B$ each agent $a \in A$ submits his or her local costs $C_a(e)$ to the mediator,%

\begin{align}
  e' = \min_{e \in B}\max_{a \in A} \{C_a(e)\}.
\end{align}

Finally, the solution $e'$ identified by Rawls voting is compared to the jointly accepted solution $e^*$ by means of complete approval voting. The winning solution becomes the new jointly accepted solution $e^*$.

\subsection{Pheromone update} \label{sec:pherup}

As the search advances, the pheromone information on each arc is continuously updated.
First, the pheromone $\tau_{(i,j)}$ of an edge $(i,j) \in E$ evaporates by the evaporation rate $\rho$.
Second, $\tau_{(i,j)}$ is increased again by $\sigma \tau^{max}$, if and only if edge $(i,j)$ is part of the path chosen by an ant (that implies $e^{*}_{(i,j)} = 1$).
Furthermore, equation (\ref{eq:pherUpdate}) takes into account the smallest and largest possible pheromone values $\tau^{min}$ and $\tau^{max}$, respectively.

\begin{align} \label{eq:pherUpdate}
\tau_{(i,t)} \leftarrow
\begin{cases}
\min\{ (1-\rho) \cdot \tau_{(i,j)} + \sigma \cdot \tau^{max} ;\: \tau^{max} \} & \text{if $e^{*}_{(i,j)} = 1$, }\\
\max \{ (1 - \rho) \cdot \tau_{(i,j)} ;\: \tau^{min} \} & \text{else.}\\
\end{cases}
\end{align}

In contrast to common approaches in the literature, the pheromone update rule (\ref{eq:pherUpdate}) does not consider the value of the objective function of a solution. In the distributed planning problem at hand, the agents do not reveal their individual costs to the mediator (cf. Section~\ref{sec:problem}) who controls the search process and updates the pheromone values. Therefore, the pheromone update depends largely on the encoded solution jointly accepted by the agents.

\section{Evaluation}
\label{sec:evaluation}

\subsection{Setup of computational study}

\subsubsection{Goals}

The metaheuristic CACM is evalauted by means of a computational benchmark study. Three goals are pursued. First, the impact of the three voting rules from Sect.~\ref{sec:voting} is studied. Second, the effect of the new collaborative local search heuristic CLS from Sect.~\ref{sec:cls} is tested. Third, the performance of CACM is compared to results from the literature, that is, to the best known solutions for the non-distributed MLULSP and other heuristics for the distributed MLULSP.

\subsubsection{Instances}

All in all, 528 benchmark instances are used in this study\footnote{Instances are available at http://www.dmlulsp.com.}. \citet{Veral_1985}, \citet{Coleman_1991}, and \citet{Dellaert_2000} introduced 176 instances for the non-distributed MLULSP.
In these instances, a single agent ($|A|=1$) is responsible to produce all items and the instances are divided into three groups of small, medium, and large instances denoted as s1, m1, and l1 (cf. Table~\ref{tab:instances}).
Based on these MLULSP instances, \citet{Homberger_2010} introduced 176 instances were the items are jointly produced by two agents ($|A|=2$) and 176 instances were the items are jointly produced by five agents ($|A|=5$).
Accordingly, these 352 DMLULSP instances are divided in six groups denoted as s2, s5, m2, m5, l2, and l5 (cf. Table~\ref{tab:instances}).
In instance group s5, each agent produces the same number of items, while in the remaining groups (s2, m2, m5, l2, and l5) this is not the case.

\begin{table}
\centering
\caption{Characteristics of used benchmark instances}
\label{tab:instances}
\vspace{5pt}
\begin{tabular}{llllll} \toprule
\multicolumn{1}{c}{class} & \multicolumn{1}{c}{group} & \multicolumn{1}{c}{$|A|$} & \multicolumn{1}{c}{$m$} & \multicolumn{1}{c}{$n$} & \multicolumn{1}{c}{no. of instances} \\ \midrule
\multicolumn{1}{c}{s} & \multicolumn{1}{c}{s1} & \multicolumn{1}{c}{1} & \multicolumn{1}{c}{5} & \multicolumn{1}{c}{12} & \multicolumn{1}{c}{96} \\
\multicolumn{1}{c}{} & \multicolumn{1}{c}{s2} & \multicolumn{1}{c}{2} & \multicolumn{1}{c}{5} & \multicolumn{1}{c}{12} & \multicolumn{1}{c}{96} \\
\multicolumn{1}{c}{} & \multicolumn{1}{c}{s5} & \multicolumn{1}{c}{5} & \multicolumn{1}{c}{5} & \multicolumn{1}{c}{12} & \multicolumn{1}{c}{96} \\[0.5em]
\multicolumn{1}{c}{m} & \multicolumn{1}{c}{m1} & \multicolumn{1}{c}{1} & \multicolumn{1}{c}{40, 50} & \multicolumn{1}{c}{12, 24} & \multicolumn{1}{c}{40} \\
\multicolumn{1}{c}{} & \multicolumn{1}{c}{m2} & \multicolumn{1}{c}{2} & \multicolumn{1}{c}{40, 50} & \multicolumn{1}{c}{12, 24} & \multicolumn{1}{c}{40} \\
\multicolumn{1}{c}{} & \multicolumn{1}{c}{m5} & \multicolumn{1}{c}{5} & \multicolumn{1}{c}{40, 50} & \multicolumn{1}{c}{12, 24} & \multicolumn{1}{c}{40} \\[0.5em]
\multicolumn{1}{c}{l} & \multicolumn{1}{c}{l1} & \multicolumn{1}{c}{1} & \multicolumn{1}{c}{500} & \multicolumn{1}{c}{36, 52} & \multicolumn{1}{c}{40} \\
\multicolumn{1}{c}{} & \multicolumn{1}{c}{l2} & \multicolumn{1}{c}{2} & \multicolumn{1}{c}{500} & \multicolumn{1}{c}{36, 52} & \multicolumn{1}{c}{40} \\
\multicolumn{1}{c}{} & \multicolumn{1}{c}{l5} & \multicolumn{1}{c}{5} & \multicolumn{1}{c}{500} & \multicolumn{1}{c}{36, 52} & \multicolumn{1}{c}{40} \\ \bottomrule
\end{tabular}
\end{table}

\subsubsection{Measure of solution quality}

In the following, the objective function value $f(y)$ (\ref{eq:global_evaluation}) of the DMLULSP is considered in the context of the non-distributed MLULSP which is an approach suggested by \citet{Dudek_2005, Dudek_2007}. For a given instance, the computed solution $y$ for the distributed MLULSP is compared to the best-known solution $y^{bk}$ for the non-distributed MLULSP and the percentage gap ${\cal G}(y)$ is calculated as

\begin{align}
{\cal G}(y) = \frac{f(y) - f^{nd}(y^{bk})}{f^{nd}(y^{bk})} 100.
\end{align}

The best-known values $f^{nd}(y^{bk})$ have been gathered from \citet{Dellaert_2000}, \citet{Pitakaso_2007}, \citet{Homberger_2008, Homberger_2010}, and \citet{Xiao_2011}.
 For the instances in s1, the optimal solutions are known.

\subsubsection{Implementation and hardware} \label{sec:eval_impl}
CACM was implemented in JAVA (JDK 1.6). The computational experiments were executed on a Linux personal computer with Intel Core 2 Duo processor 1.83 GHz. To compare the heuristics in Sect.~\ref{sec:eval_comp}, all experiments were executed on this computer.

\subsubsection{Parameters}

By means of a preliminary study, the parameters of CACM shown in Table~\ref{tab:parameter} have been determined. For all instances, the pheromone evaporation rate $\rho$, the pheromone intensification factor $\sigma$, and the minimum pheromone value $\tau^{min}$ per edge are constant. Then again, the number of generated solutions $\overline{s}$ and the maximum pheromone value $\tau^{max}$ per edge grow with increasing instance size. The ballot size $b$ is evaluated together with the voting rules in the next section.

\begin{table}
\centering
\caption{Parameter during evaluation of metaheuristic CACM}
\label{tab:parameter}
\vspace{5pt}
  \begin{tabular}{llll} \toprule
     & \multicolumn{1}{c}{class s} & \multicolumn{1}{c}{class m} & \multicolumn{1}{c}{class l} \\ \midrule
    $\tau^{min}$ & \multicolumn{1}{r}{1} & \multicolumn{1}{r}{1} & \multicolumn{1}{r}{1} \\
    $\tau^{max}$ & \multicolumn{1}{r}{100} & \multicolumn{1}{r}{1,000} & \multicolumn{1}{r}{1,500} \\
    $\rho$ & \multicolumn{1}{r}{5\%} & \multicolumn{1}{r}{5\%} & \multicolumn{1}{r}{5\%} \\
    $\sigma$ & \multicolumn{1}{r}{5\%} & \multicolumn{1}{r}{5\%} & \multicolumn{1}{r}{5\%} \\
    $\overline{s}$ & \multicolumn{1}{r}{50,000} & \multicolumn{1}{r}{200,000} & \multicolumn{1}{r}{400,000} \\ \bottomrule
  \end{tabular}
\end{table}

\subsection{Effect of voting rules and ballot size}
The effects of the Borda and Rawls voting rules on the performance of CACM are studied by means of the benchmark instances from groups s2 and s5.
Both voting rules are applied using ballot sizes $b$ of 1, 10, 20, and 100. The results are given in Table~\ref{tab:votingrules}.

\begin{table}
\centering
\caption{Effect of Borda and Rawls voting}
\label{tab:votingrules}
\vspace{5pt}

    \begin{tabular}{llll} \toprule
    Agents & \multicolumn{1}{c}{Ballot size $b$} & \multicolumn{2}{c}{${\cal G}$} \\ \cmidrule(lr){3-4}
     & \multicolumn{1}{c}{} & \multicolumn{1}{c}{Borda} & \multicolumn{1}{c}{Rawls} \\ \midrule
    $|A|=2$ & \multicolumn{1}{r}{1} & \multicolumn{1}{r}{2.21} & \multicolumn{1}{r}{2.21} \\
     & \multicolumn{1}{r}{10} & \multicolumn{1}{r}{2.14} & \multicolumn{1}{r}{2.13} \\
     & \multicolumn{1}{r}{20} & \multicolumn{1}{r}{2.40} & \multicolumn{1}{r}{2.31} \\
     & \multicolumn{1}{r}{100} & \multicolumn{1}{r}{2.46} & \multicolumn{1}{r}{2.55} \\ [0.5em]
    $|A|=5$ & \multicolumn{1}{r}{1} & \multicolumn{1}{r}{8.31} & \multicolumn{1}{r}{8.31} \\
     & \multicolumn{1}{r}{10} & \multicolumn{1}{r}{8.78} & \multicolumn{1}{r}{8.98} \\
     & \multicolumn{1}{r}{20} & \multicolumn{1}{r}{8.62} & \multicolumn{1}{r}{9.21} \\
     & \multicolumn{1}{r}{100} & \multicolumn{1}{r}{9.22} & \multicolumn{1}{r}{9.41} \\ \bottomrule
    \end{tabular}
\end{table}

The configuration of Borda voting and Rawls voting with a ballot size of $b=1$ results in equivalent voting procedures and, therefore, leads to identical results (see Table~\ref{tab:votingrules}).
Especially in case of scenarios with five agents, the results for the Borda voting rule are slightly better than the results for the Rawls voting rule.
However, in the five agents scenario, both voting rules lead to the best results for ballot sizes of only one. For the two agent scenario, ballot sizes of one lead to the second best results.
All in all, both voting rules seem to result in higher total costs compared to the degenerated case ($b=1$).
Therefore, the parameter ballot size $b$ is set to one, that is, the function 'voting($B,A$)' in Alg.~\ref{alg:caco} is in fact deactivated. The only voting mechanism applied is complete approval voting which is used during collaborative local search. A positive side effect of doing without Rawls and Borda voting is that no local cost information have to be revealed (Rawls) and the opportunity of strategic bidding, which is intrinsic in ranked based voting, is reduced (Borda).

\subsection{Effect of collaborative local search}

In this test, the effect of CLS is studied. The CACM approach including collaborative local search (CLS, cf. Sect.~\ref{sec:cls}) is compared with CACM excluding CLS. The latter is denoted as CACM$^{ex}$. As described previously, only complete approval voting is used. Furthermore, CLS is used as stand-alone heuristic which follows Alg.~\ref{alg:caco} without the 'constructEncodedSolution' procedure. The initial encoded solution $e$ is set to the unit matrix.

The three variants are compared by means of the instances from groups s1, s2, and s5.
The instances of s1 correspond to the non-distributed MLULSP for which optimal solutions are known.
Consequently, Table~\ref{tab:heuristic} depicts the average gap $\cal G$ with respect to the optimal MLULSP solution.

\begin{table}
\centering
\caption{Effect of collaborative local search.}
\label{tab:heuristic}
\vspace{5pt}
  \begin{tabular}{lrrr} \toprule
  Heuristic & \multicolumn{3}{c}{${\cal G}$} \\ \cmidrule(lr){2-4}
     & \multicolumn{1}{c}{s1} & \multicolumn{1}{c}{s2} & \multicolumn{1}{c}{s5} \\ \midrule
    CLS & 3.93 & 5.29 & 9.53 \\
    CACM$^{ex}$ & 1.65 & 4.96 & 9.15 \\
    CACM & \emph{0.07} & \emph{2.21} & \emph{8.31} \\ \bottomrule
  \end{tabular}
\end{table}

\begin{figure}
\vspace{36pt}
\centering
  \includegraphics[width=11cm]{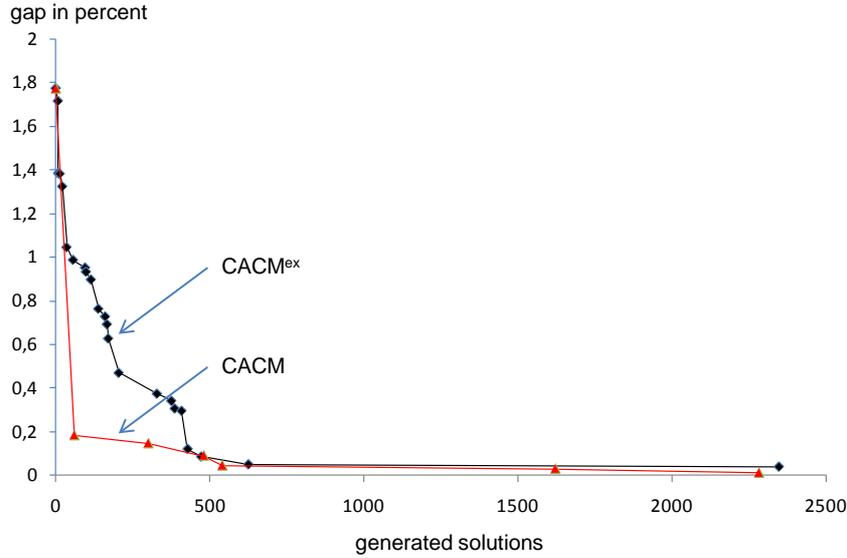}\\
  \caption{Convergency of CACM and CACM$^{ex}$.}\label{fig:convergence}
\end{figure}

CACM solves 80 of 96 instances to optimality. For the remaining 16 instances, the average optimality gap is only 0.07 percent although CACM does not use heuristic information specific to the MLULSP to control the search process.
Furthermore, CACM outperforms the stand-alone CLS and CACM$^{ex}$ both in the non-distributed (s1) and distributed (s2, s5) scenario. Obviously, the combination of CLS and CACM$^{ex}$ increases the overall performance. Finally, Figure~\ref{fig:convergence} shows exemplarily for an instance, that the heuristic CACM converges faster than the heuristic CACM$^{ex}$.

\subsection{Comparison with heuristics from the literature}
\label{sec:eval_comp}

CACM is compared to three approaches designed for the DMLUSLP denoted here as SA, ES, and ACO which are based on the metaheuristics simulated annealing, evolutionary strategy, and ant colony optimization, respectively. SA was introduced by \citep{Homberger_2010},  ES is presented in \citet{Homberger_2011}, and ACO is discussed in \citep{Homberger_2010a, Homberger_2011a}. Some of these previous computational experiments were repeated or broadened to take into account additional instances (see Section~\ref{sec:eval_impl}).

Table~\ref{tab:comparison} shows the average gap $\cal G$ over 176 instances with two agents (classes s2,m2,l2) as well as over 176 instances with five agents (classes s5, m5, l5). The detailed results for medium and large instances are given in Tables \ref{tab:res_m2}, \ref{tab:res_m5}, \ref{tab:res_l2}, and \ref{tab:res_l5} in the appendix.
Due to long runtimes \citet{Homberger_2010a, Homberger_2011a} compared their approach only by means of the medium sized instances of groups m2 and m5.

\begin{table}
\centering
\caption{Comparison of different heuristics by means of average percentage gap $\cal G$ from best-known non-distributed solutions.}
\label{tab:comparison}
\vspace{5pt}
\scriptsize
    \begin{tabular}{lllllllllllll} \toprule
     & \multicolumn{4}{c}{Class s} & \multicolumn{4}{c}{Class m} & \multicolumn{4}{c}{Class l} \\ \cmidrule(lr){2-5} \cmidrule(lr){6-9} \cmidrule(lr){10-13}
     & \multicolumn{1}{c}{ES} & \multicolumn{1}{c}{SA} & \multicolumn{1}{c}{ACO} & \multicolumn{1}{c}{CACM} & \multicolumn{1}{c}{ES} & \multicolumn{1}{c}{SA} & \multicolumn{1}{c}{ACO} & \multicolumn{1}{c}{CACM} & \multicolumn{1}{c}{ES} & \multicolumn{1}{c}{SA} & \multicolumn{1}{c}{ACO} & \multicolumn{1}{c}{CACM} \\ \midrule
    $|A|=2$ & \multicolumn{1}{r}{5.92} & \multicolumn{1}{r}{\emph{1.78}} & \multicolumn{1}{r}{2.66} & \multicolumn{1}{r}{2.21} & \multicolumn{1}{r}{18.65} & \multicolumn{1}{r}{7.95} & \multicolumn{1}{r}{2.77} & \multicolumn{1}{r}{\emph{1.84}} & \multicolumn{1}{r}{55.07} & \multicolumn{1}{r}{16.84} & \multicolumn{1}{r}{13.30} & \multicolumn{1}{r}{\emph{5.43}} \\
    $|A|=5$ & \multicolumn{1}{r}{18.66} & \multicolumn{1}{r}{\emph{3.18}} & \multicolumn{1}{r}{7.63} & \multicolumn{1}{r}{8.31} & \multicolumn{1}{r}{55.38} & \multicolumn{1}{r}{16.55} & \multicolumn{1}{r}{8.18} & \multicolumn{1}{r}{\emph{7.74}} & \multicolumn{1}{r}{95.93} & \multicolumn{1}{r}{46.38} & \multicolumn{1}{r}{23.79} & \multicolumn{1}{r}{\emph{5.17}} \\ \bottomrule
    \end{tabular}
\end{table}

The presented CACM approach significantly improves the best known solutions for the medium and large instances (see Table~\ref{tab:comparison}).
For example, CACM reduces the average gap $\cal G$ of the large instances with five agents (group l5) from 46 percent to only about 5 percent.
In detail, CACM improves twenty-nine out of forty m2 instances, twenty out of forty m5 instances, thirty-six out of forty l2 instances, and forty out of forty l5 instances (see appendix, Tables \ref{tab:res_m2}, \ref{tab:res_m5}, \ref{tab:res_l2}, and \ref{tab:res_l5}).
With respect to the classes of small instances s2 and s5, CACM is the second best procedure.

\section{Conclusions}
\label{sec:conclusion}

A collaborative solution approach based on the metaheuristic ant colony optimization was presented. It is used to coordinate joint production planning of a coalition of self-interested decision makers with asymmetric information. The search process is executed by a mediator who is not aware of the agents local costs. Furthermore, the agents do not have to reveal private cost information during search. Nevertheless, to control the search process, the mediator tries to elicit preferences by means of voting and a new collaborative local search procedure. The results are used to update the pheromone information which guide the solution construction process. Solutions are constructed by means of a new graph structure based on encoded solutions which is less complicated and therefore easier to search. The new approach CACM is evaluated by means of 356 benchmark instances from the literature and outperforms all existing distributed planning approaches on the sets of medium and large instances.

For future research, it appears interesting to study collaborative lot sizing problems with a non-disjunct assignment of items to agents. This would increase potential planning conflicts between the agents and opens up interesting challenges for developing collaborative planning metaheuristics.

\bibliographystyle{chicago}

\begin{thebibliography}{012345678901234567890123456789}

\bibitem[\protect\citeauthoryear{Arkin, Joneja, and Roundy}{Arkin
  et~al.}{1989}]{Arkin_1989}
Arkin, E., D.~Joneja, and R.~Roundy (1989).
\newblock Computational complexity of uncapacitated multi-echelon production
  planning problems.
\newblock {\em Operations Research Letters\/}~{\em 8\/}(2), 61--66.

\bibitem[\protect\citeauthoryear{Bookbinder and Koch}{Bookbinder and
  Koch}{1990}]{Bookbinder_1990}
Bookbinder, J. and L.~Koch (1990).
\newblock Production planning for mixed assembly/arborescent systems.
\newblock {\em Journal of Operations Management\/}~{\em 9}, 7--23.

\bibitem[\protect\citeauthoryear{Brams and Fishburn}{Brams and
  Fishburn}{1978}]{Brams_1978}
Brams, S.~J. and P.~C. Fishburn (1978).
\newblock Approval voting.
\newblock {\em The American Political Science Review\/}~{\em 72\/}(3), pp.
  831--847.

\bibitem[\protect\citeauthoryear{Coleman and McKnew}{Coleman and
  McKnew}{1991}]{Coleman_1991}
Coleman, B. and M.~McKnew (1991).
\newblock An improved heuristic for multilevel lot sizing in material
  requirements planning.
\newblock {\em Decision Sciences\/}~{\em 22\/}(1), 136--156.

\bibitem[\protect\citeauthoryear{Dellaert and Jeunet}{Dellaert and
  Jeunet}{2000}]{Dellaert_2000}
Dellaert, N. and J.~Jeunet (2000).
\newblock Solving large unconstrained multilevel lot-sizing problems using a
  hybrid genetic algorithm.
\newblock {\em International Journal of Production Research\/}~{\em 38\/}(5),
  1083--1099.

\bibitem[\protect\citeauthoryear{Dellaert and Jeunet}{Dellaert and
  Jeunet}{2003}]{Dellaert_2003}
Dellaert, N. and J.~Jeunet (2003).
\newblock Randomized multi-level lot-sizing heuristics for general product
  structures.
\newblock {\em European Journal of Operational Research\/}~{\em 148\/}(1),
  211--228.

\bibitem[\protect\citeauthoryear{Drechsel and Kimms}{Drechsel and
  Kimms}{2010}]{Drechsel_2010}
Drechsel, J. and A.~Kimms (2010).
\newblock The subcoalition-perfect core of cooperative games.
\newblock {\em Annals of Operations Research\/}~{\em 181}, 591--601.
\newblock 10.1007/s10479-010-0789-8.

\bibitem[\protect\citeauthoryear{Drechsel and Kimms}{Drechsel and
  Kimms}{2011}]{Drechsel_2011}
Drechsel, J. and A.~Kimms (2011).
\newblock Cooperative lot sizing with transshipments and scarce capacities:
  solutions and fair cost allocations.
\newblock {\em International Journal of Production Research\/}~{\em 49\/}(9),
  2643--2668.

\bibitem[\protect\citeauthoryear{Drexl and Kimms}{Drexl and
  Kimms}{1997}]{Drexl_1997}
Drexl, A. and A.~Kimms (1997).
\newblock Lot sizing and scheduling -- survey and extensions.
\newblock {\em European Journal of Operational Research\/}~{\em 99\/}(2), 221
  -- 235.

\bibitem[\protect\citeauthoryear{Dudek and Stadtler}{Dudek and
  Stadtler}{2005}]{Dudek_2005}
Dudek, G. and H.~Stadtler (2005).
\newblock Negotiation-based collaborative planning between supply chains
  partners.
\newblock {\em European Journal of Operational Research\/}~{\em 163\/}(3),
  668--687.

\bibitem[\protect\citeauthoryear{Dudek and Stadtler}{Dudek and
  Stadtler}{2007}]{Dudek_2007}
Dudek, G. and H.~Stadtler (2007).
\newblock Negotiation-based collaborative planning in divergent two-tier supply
  chains.
\newblock {\em International Journal of Production Research\/}~{\em 45\/}(2),
  465--484.

\bibitem[\protect\citeauthoryear{Ertogral and Wu}{Ertogral and
  Wu}{2000}]{Ertogral_2000}
Ertogral, K. and S.~Wu (2000).
\newblock Auction-theoretic coordination of production planning in the supply
  chain.
\newblock {\em IIE Transactions\/}~{\em 32\/}(10), 931--940.

\bibitem[\protect\citeauthoryear{Fink}{Fink}{2004}]{Fink_2004a}
Fink, A. (2004).
\newblock Supply chain coordination by means of automated negotiations.
\newblock In {\em In Proceedings of the 37th Hawaii International Conference on
  System Sciences}.

\bibitem[\protect\citeauthoryear{Fink}{Fink}{2006}]{Fink_2006}
Fink, A. (2006).
\newblock {\em Multiagent-based supply chain management}, Chapter Supply chain
  coordination by means of automated negotiation between autonomous agents,
  pp.\  351--372.
\newblock Berlin: Springer.

\bibitem[\protect\citeauthoryear{Fink}{Fink}{2007}]{Fink_2007}
Fink, A. (2007).
\newblock Barwertorientierte projektplanung mit mehreren akteuren mittels eines
  verhandlungsbasierten koordinationsmechanismus.
\newblock In O.~Oberweis, C.~Weinhardt, H.~Gimpel, A.~Koschmider,
  V.~Pankratius, and B.~Schnizler (Eds.), {\em eOrganisation: Service-,
  Prozess-, Market-Engineering: 8. Internationale Tagung Wirtschaftsinformatik
  2007}, pp.\  465--482. Karlsruhe: Universit\"{a}tsverlag.

\bibitem[\protect\citeauthoryear{Frayret}{Frayret}{2009}]{Frayret_2009}
Frayret, J.-M. (2009, oct.).
\newblock A multidisciplinary review of collaborative supply chain planning.
\newblock In {\em Systems, Man and Cybernetics, 2009. SMC 2009. IEEE
  International Conference on}, pp.\  4414 --4421.

\bibitem[\protect\citeauthoryear{Homberger}{Homberger}{2008}]{Homberger_2008}
Homberger, J. (2008).
\newblock A parallel genetic algorithm for the multi-level unconstrained
  lot-sizing problem.
\newblock {\em INFORMS Journal on Computing\/}~{\em 20\/}(1), 124--132.

\bibitem[\protect\citeauthoryear{Homberger}{Homberger}{2010}]{Homberger_2010}
Homberger, J. (2010).
\newblock Decentralized multi-level uncapacitated lot-sizing by automated
  negotiation.
\newblock {\em 4OR: A Quarterly Journal of Operations Research\/}~{\em 8},
  155--180.

\bibitem[\protect\citeauthoryear{Homberger}{Homberger}{2011}]{Homberger_2011}
Homberger, J. (2011).
\newblock A generic coordination mechanism for lot-sizing in supply chains.
\newblock {\em Electronic Commerce Research\/}~{\em 11\/}(2), 123--149.

\bibitem[\protect\citeauthoryear{Homberger and Gehring}{Homberger and
  Gehring}{2009}]{Homberger_2009a}
Homberger, J. and H.~Gehring (2009).
\newblock An ant colony optimization approach for the multi-level unconstrained
  lot-sizing problem.
\newblock In {\em Proceedings of the 42nd Hawaii International Conference on
  System Sciences}, Washington, DC, pp.\  1--7. IEEE Computer Society.

\bibitem[\protect\citeauthoryear{Homberger and Gehring}{Homberger and
  Gehring}{2010}]{Homberger_2010a}
Homberger, J. and H.~Gehring (2010).
\newblock A pheromone-based negotiation mechanism for lot-sizing in supply
  chains.
\newblock In {\em Proceedings of the 44nd Hawaii International Conference on
  System Sciences}, pp.\  1--10.

\bibitem[\protect\citeauthoryear{Homberger and Gehring}{Homberger and
  Gehring}{2011}]{Homberger_2011a}
Homberger, J. and H.~Gehring (2011).
\newblock An ant colony optimization-based negotation approach for lot-sizing
  in supply chains.
\newblock {\em International Journal of Information Processing and
  Management\/}~{\em 2\/}(3), 86--99.

\bibitem[\protect\citeauthoryear{Jans and Degraeve}{Jans and
  Degraeve}{2007}]{Jans_2007}
Jans, R. and Z.~Degraeve (2007).
\newblock Meta-heuristics for dynamic lot sizing: A review and comparison of
  solution approaches.
\newblock {\em European Journal of Operational Research\/}~{\em 177\/}(3), 1855
  -- 1875.

\bibitem[\protect\citeauthoryear{Jung, Chen, and Jeong}{Jung
  et~al.}{2005}]{Jung_2005}
Jung, H., F.~Chen, and B.~Jeong (2005).
\newblock A production-distribution coordinating model for third party
  logistics partnership.
\newblock In {\em Proc. of the 2005 IEEE Internat. Conf. on Autom. Sci. and
  Eng.}, Edmonton, Canada, pp.\  99--104.

\bibitem[\protect\citeauthoryear{Lee and Kumara}{Lee and
  Kumara}{2007a}]{Lee_2007b}
Lee, S. and S.~Kumara (2007a).
\newblock Decentralized supply chain coordination through auction markets:
  dynamic lot-sizing in distribution networks.
\newblock {\em International Journal of Production Research\/}~{\em 45\/}(20),
  4715--4733.

\bibitem[\protect\citeauthoryear{Lee and Kumara}{Lee and
  Kumara}{2007b}]{Lee_2007}
Lee, S. and S.~Kumara (2007b).
\newblock {\em Multiagent system approach for dynamic lot-sizing in supply
  chains}, Chapter Trends in supply chain design and management - part II, pp.\
   311--330.
\newblock London: Springer.

\bibitem[\protect\citeauthoryear{Lomuscio, Wooldridge, and Jennings}{Lomuscio
  et~al.}{2003}]{Lomuscio_2003}
Lomuscio, A., M.~Wooldridge, and N.~Jennings (2003).
\newblock A classification scheme for negotiation in electronic commerce.
\newblock {\em Group Decision and Negotiation\/}~{\em 12\/}(1), 31--56.

\bibitem[\protect\citeauthoryear{Pitakaso, Almeder, Doerner, and
  Hartl}{Pitakaso et~al.}{2007}]{Pitakaso_2007}
Pitakaso, R., C.~Almeder, K.~Doerner, and R.~Hartl (2007).
\newblock A max-min ant system for unconstrained multi-level lot-sizing
  problems.
\newblock {\em Computers \& Operations Research\/}~{\em 34\/}(9), 2533--2552.

\bibitem[\protect\citeauthoryear{Rawls}{Rawls}{1971}]{Rawls_1971}
Rawls, J. (1971).
\newblock {\em A theory of justice}.
\newblock Oxford: Oxford University Press.

\bibitem[\protect\citeauthoryear{Salomon and Kuik}{Salomon and
  Kuik}{1993}]{Salomon_1993}
Salomon, M. and R.~Kuik (1993).
\newblock Statistical search methods for lotsizing problems.
\newblock {\em Annals of Operations Research\/}~{\em 41\/}(4), 453--468.

\bibitem[\protect\citeauthoryear{Stadtler}{Stadtler}{2009}]{Stadtler_2009}
Stadtler, H. (2009).
\newblock A framework for collaborative planning and state-of-the-art.
\newblock {\em OR Spectrum\/}~{\em 31\/}(1), 5--30.

\bibitem[\protect\citeauthoryear{Steinberg and Napier}{Steinberg and
  Napier}{1980}]{Steinberg_1980}
Steinberg, E. and H.~Napier (1980).
\newblock Optimal multi-level lot sizing for requirements planning systems.
\newblock {\em Management Science\/}~{\em 26\/}(12), 1258--1271.

\bibitem[\protect\citeauthoryear{Str\"{o}bel and Weinhardt}{Str\"{o}bel and
  Weinhardt}{2003}]{Stroebel_2003}
Str\"{o}bel, M. and C.~Weinhardt (2003).
\newblock The montreal taxonomy for electronic negotiations.
\newblock {\em Group Decision and Negotiation\/}~{\em 12\/}(2), 143--164.

\bibitem[\protect\citeauthoryear{St\"{u}tzle and Hoos}{St\"{u}tzle and
  Hoos}{2000}]{Stuetzle_2000}
St\"{u}tzle, T. and H.~Hoos (2000).
\newblock Max-min ant system.
\newblock {\em Future Generation Computer Systems\/}~{\em 16\/}(9), 889--914.

\bibitem[\protect\citeauthoryear{Veral and LaForge}{Veral and
  LaForge}{1985}]{Veral_1985}
Veral, E. and R.~LaForge (1985).
\newblock The performance of a simple incremental lot-sizing rule in a
  multilevel inventory environment.
\newblock {\em Decision Sciences\/}~{\em 16\/}(1), 57--72.

\bibitem[\protect\citeauthoryear{Vo\ss and Woodruff}{Vo\ss and
  Woodruff}{2006}]{Voss_2006}
Vo\ss, S. and D.~Woodruff (2006).
\newblock {\em Introduction to computational optimization models for production
  planning in a supply chain\/} (2nd ed.), Volume Berlin.
\newblock Springer.

\bibitem[\protect\citeauthoryear{Winands, Adan, and van Houtum}{Winands
  et~al.}{2011}]{Winands_2011}
Winands, E., I.~Adan, and G.~van Houtum (2011).
\newblock The stochastic economic lot scheduling problem: A survey.
\newblock {\em European Journal of Operational Research\/}~{\em 210\/}(1), 1 --
  9.

\bibitem[\protect\citeauthoryear{Xiao, Kaku, Zhao, and Zhang}{Xiao
  et~al.}{2011}]{Xiao_2011}
Xiao, Y., I.~Kaku, Q.~Zhao, and R.~Zhang (2011).
\newblock A reduced variable neighborhood search algorithm for uncapacitated
  multilevel lot-sizing problems.
\newblock {\em European Journal of Operational Research\/}~{\em 214\/}(2), 223
  -- 231.

\bibitem[\protect\citeauthoryear{Xiao, Kaku, Zhao, and Zhang}{Xiao
  et~al.}{2012}]{Xiao_2012}
Xiao, Y., I.~Kaku, Q.~Zhao, and R.~Zhang (2012).
\newblock Neighborhood search techniques for solving uncapacitated multilevel
  lot-sizing problems.
\newblock {\em Computers \& Operations Research\/}~{\em 39\/}(3), 647 -- 658.

\bibitem[\protect\citeauthoryear{Yelle}{Yelle}{1979}]{Yelle_1979}
Yelle, L. (1979).
\newblock Materials requirements lot sizing: A multilevel approach.
\newblock {\em International Journal of Production Research\/}~{\em 17\/}(3),
  223--232.

\bibitem[\protect\citeauthoryear{Zimmer}{Zimmer}{2004}]{Zimmer_2004}
Zimmer, K. (2004).
\newblock Hierarchical coordination mechanisms within the supply chain.
\newblock {\em European Journal of Operational Research\/}~{\em 153\/}(3),
  687--703.

\end{thebibliography}

\clearpage
\newgeometry{a4paper, top=18mm, left=18mm, right=18mm, bottom=10mm, footskip=0mm, headsep=0mm, headheight=0mm}
\section*{Appendix: Detailed Computational Results}

\begin{table}[h]
\centering
\caption{Results per instance of class $m$ with two agents ($|A|=2$)}{
\label{tab:res_m2} \vspace{1pt}
\footnotesize
\begin{tabular}{rrrrrr} \toprule
 & \multicolumn{1}{c}{$|A| = 1$} & \multicolumn{4}{c}{distributed to $|A| = 2$ agents} \\ \cmidrule(lr){2-2} \cmidrule(lr){3-6}
\multicolumn{1}{l}{class $m$ instance} & \multicolumn{1}{c}{best-known} & \multicolumn{1}{c}{ES} & \multicolumn{1}{c}{SA} & \multicolumn{1}{c}{ACO} & \multicolumn{1}{c}{CACM} \\ \midrule
1 & 194,571.00 & 226,310.55 & 199,961.45 & 202,824.35 & \textbf{198,913.65}  \\
2 & 165,110.00 & 189,359.10 & 172,215.95 & \textbf{168,960.35} & 171,589.85  \\
3 & 201,226.00 & 243,635.40 & 207,379.75 & 207,244.65 & \textbf{201,436.75}  \\
4 & 187,790.00 & 224,011.65 & 191,731.45 & 191,036.65 & \textbf{189,687.65}  \\
5 & 161,304.00 & 184,229.85 & 165,815.30 & 166,007.00 & \textbf{164,463.90}  \\
6 & 342,916.00 & 397,948.15 & 351,960.65 & \textbf{349,865.05} & 351,508.00  \\
7 & 292,908.00 & 338,136.05 & 305,473.45 & 299,409.35 & \textbf{297,715.40}  \\
8 & 354,919.00 & 422,567.70 & 364,522.50 & 367,615.55 & \textbf{362,346.40}  \\
9 & 325,212.00 & 357,269.80 & 333,153.40 & 334,031.75 & \textbf{332,133.80}  \\
10 & 385,939.00 & 441,387.35 & 407,565.35 & 398,898.65 & \textbf{393,280.40}  \\
11 & 179,762.00 & 192,527.35 & 189,529.80 & 204,292.60 & \textbf{179,839.00}  \\
12 & 155,938.00 & 164,848.85 & 182,688.95 & 157,301.95 & \textbf{156,130.95}  \\
13 & 183,219.00 & 191,549.55 & 189,108.80 & 188,364.60 & \textbf{183,242.25}  \\
14 & 136,462.00 & 145,731.45 & 140,651.90 & \textbf{138,738.35} & 141,477.60  \\
15 & 186,597.00 & 214,290.70 & 191,057.20 & 190,761.45 & \textbf{188,255.45}  \\
16 & 340,686.00 & 397,568.85 & 351,210.65 & 357,909.70 & \textbf{349,256.35}  \\
17 & 378,845.00 & 448,356.00 & 393,817.25 & 391,014.60 & \textbf{382,513.70}  \\
18 & 346,313.00 & 411,172.95 & 372,168.80 & 357,764.35 & \textbf{356,937.55}  \\
19 & 411,997.00 & 464,478.90 & 442,328.20 & 421,823.70 & \textbf{417,107.60}  \\
20 & 390,194.00 & 426,386.00 & 412,580.70 & \textbf{402,375.25} & 407,367.20  \\
21 & 148,004.00 & 214,663.10 & 170,440.75 & 148,628.90 & \textbf{148,083.70}  \\
22 & 197,695.00 & 248,894.10 & 241,675.30 & 201,297.00 & \textbf{200,501.70}  \\
23 & 160,693.00 & 201,806.10 & 178,578.30 & \textbf{160,924.90} & \textbf{160,924.90}  \\
24 & 184,358.00 & 216,154.35 & 220,872.40 & 185,426.70 & \textbf{185,157.20}  \\
25 & 161,457.00 & 205,387.90 & 177,611.55 & 163,421.00 & \textbf{161,967.00}  \\
26 & 344,970.00 & 448,962.10 & 404,290.70 & \textbf{349,062.25} & 349,649.20  \\
27 & 352,634.00 & 471,776.05 & 403,475.40 & \textbf{363,633.60} & 369,683.25  \\
28 & 356,323.00 & 480,489.55 & 435,463.50 & 363,952.10 & \textbf{361,585.10}  \\
29 & 411,338.00 & 651,772.35 & 499,775.40 & 431,998.00 & \textbf{429,759.90}  \\
30 & 401,732.00 & 521,972.05 & 442,778.65 & \textbf{414,748.15} & 414,996.80  \\
31 & 185,161.00 & 210,866.20 & 200,677.75 & \textbf{189,066.75} & 191,642.55  \\
32 & 185,542.00 & 207,226.15 & 193,634.60 & 191,893.95 & \textbf{188,634.85}  \\
33 & 192,157.00 & 211,073.20 & 204,163.95 & 201,108.45 & \textbf{195,519.95}  \\
34 & 136,757.00 & 152,061.20 & 147,227.50 & \textbf{138,928.40} & 138,987.85  \\
35 & 166,041.00 & 181,383.50 & 177,690.45 & 170,896.25 & \textbf{168,874.65}  \\
36 & 289,846.00 & 332,234.45 & 308,642.60 & \textbf{292,530.95} & 293,472.75  \\
37 & 337,913.00 & 384,122.10 & 354,226.60 & 348,464.80 & \textbf{344,700.05}  \\
38 & 319,905.00 & 377,400.80 & 342,019.30 & \textbf{327,445.40} & 328,214.05  \\
39 & 366,848.00 & 424,664.05 & 387,213.45 & 377,137.50 & \textbf{374,957.45}  \\
40 & 305,011.00 & 380,260.00 & 319,977.20 & 310,249.80 & \textbf{308,190.20}  \\ \midrule
median & 245,536.00 & 290,564.28 & 273,574.38 & 249,887.80 & 247,454.75  \\
mean & 263,157.33 & 315,123.39 & 284,383.92 & 270,676.37 & 268,517.66  \\
stand. dev. & 94,057.58 & 125,416.02 & 104,568.95 & 97,388.37 & 97,485.57  \\ [0.2em]
median $\cal G$ &  & 15.90 & 5.67 & 2.55 & 1.69   \\
mean $\cal G$ &  & 18.65 & 7.95 & 2.77 & 1.84   \\
stand. dev.  $\cal G$ &  & 10.64 & 5.96 & 2.09 & 1.28   \\ [0.2em]
no. of best known$^{\rm a}$  & & 0 & 0 & 11 & 30   \\ \bottomrule
\end{tabular}}
\vspace{1pt}
\scriptsize

$^{\rm a}$Both ACO and CACM compute the best known solution for instance 23.
\end{table}

\begin{table}
\centering
\caption{Results per instance of class $m$ with five agents ($|A|=5$)}
\label{tab:res_m5}
\vspace{5pt}
\footnotesize
\begin{tabular}{rrrrrr} \toprule
 & \multicolumn{1}{c}{$|A| = 1$} & \multicolumn{4}{c}{distributed to $|A| = 5$ agents} \\ \cmidrule(lr){2-2} \cmidrule(lr){3-6}
\multicolumn{1}{l}{class $m$ instance} & \multicolumn{1}{c}{best-known} & \multicolumn{1}{c}{ES} & \multicolumn{1}{c}{SA} & \multicolumn{1}{c}{ACO} & \multicolumn{1}{c}{CACM} \\ \midrule
1 & 194,571.00 & 264,065.00 & 205,787.15 & \textbf{204,753.85} & 207,630.80  \\
2 & 165,110.00 & 231,090.80 & 173,471.95 & \textbf{173,361.15} & 174,708.10  \\
3 & 201,226.00 & 280,535.50 & 226,796.85 & \textbf{214,696.05} & 214,744.65  \\
4 & 187,790.00 & 273,135.80 & \textbf{196,780.85} & 205,980.65 & 203,166.05  \\
5 & 161,304.00 & 200,282.10 & 173,262.45 & 174,641.15 & \textbf{168,236.50}  \\
6 & 342,916.00 & 493,419.20 & \textbf{368,168.85} & 369,897.35 & 370,721.90  \\
7 & 292,908.00 & 403,347.70 & 320,741.40 & 316,507.00 & \textbf{313,263.90}  \\
8 & 354,919.00 & 563,164.80 & 376,777.75 & \textbf{373,085.55} & 373,117.70  \\
9 & 325,212.00 & 447,899.60 & \textbf{345,253.25} & 359,006.95 & 359,566.55  \\
10 & 385,939.00 & 551,828.05 & 441,193.25 & \textbf{425,803.55} & 429,034.55  \\
11 & 179,762.00 & 312,657.85 & 215,994.40 & 200,473.80 & \textbf{197,660.40}  \\
12 & 155,938.00 & 198,585.80 & 173,838.45 & 171,895.75 & \textbf{168,395.75}  \\
13 & 183,219.00 & 251,756.45 & 227,887.45 & 207,910.65 & \textbf{202,287.10}  \\
14 & 136,462.00 & 165,146.70 & \textbf{142,493.75} & 148,623.95 & 153,552.15  \\
15 & 186,597.00 & 343,134.10 & \textbf{193,885.00} & 203,231.10 & 198,769.05  \\
16 & 340,686.00 & 541,487.35 & \textbf{377,353.90} & 377,807.85 & 377,726.10  \\
17 & 378,845.00 & 596,551.85 & 458,401.05 & \textbf{413,040.65} & 414,022.55  \\
18 & 346,313.00 & 525,989.70 & 405,675.05 & 391,587.65 & \textbf{390,264.35}  \\
19 & 411,997.00 & 559,655.00 & \textbf{456,057.80} & 468,349.70 & 468,899.25  \\
20 & 390,194.00 & 600,109.20 & 480,153.95 & 457,261.15 & \textbf{449,742.00}  \\
21 & 148,004.00 & 211,273.15 & 186,605.50 & 154,271.80 & \textbf{153,865.00}  \\
22 & 197,695.00 & 324,509.20 & 239,446.35 & 205,484.50 & \textbf{204,578.80}  \\
23 & 160,693.00 & 261,299.90 & 178,689.70 & 171,443.70 & \textbf{168,981.65}  \\
24 & 184,358.00 & 303,488.65 & 226,545.15 & 190,604.80 & \textbf{188,964.45}  \\
25 & 161,457.00 & 248,752.95 & 210,579.05 & 169,652.40 & \textbf{168,805.40}  \\
26 & 344,970.00 & 775,580.95 & 434,723.60 & 376,325.90 & \textbf{368,714.30}  \\
27 & 352,634.00 & 768,345.20 & 596,826.80 & 385,677.45 & \textbf{383,341.15}  \\
28 & 356,323.00 & 803,754.00 & 464,721.50 & \textbf{380,075.70} & 383,552.95  \\
29 & 411,338.00 & 845,876.45 & 718,175.10 & 459,726.25 & \textbf{459,557.10}  \\
30 & 401,732.00 & 833,605.10 & 458,301.15 & 432,181.35 & \textbf{429,949.70}  \\
31 & 185,161.00 & 249,657.65 & 207,223.10 & 201,043.10 & \textbf{198,819.55}  \\
32 & 185,542.00 & 262,525.70 & 205,138.90 & \textbf{195,533.20} & 197,363.30  \\
33 & 192,157.00 & 258,188.80 & 211,960.20 & 205,325.50 & \textbf{203,314.05}  \\
34 & 136,757.00 & 210,986.15 & 155,276.20 & \textbf{141,500.90} & 142,881.25  \\
35 & 166,041.00 & 238,389.35 & \textbf{178,082.85} & 179,720.25 & 181,454.00  \\
36 & 289,846.00 & 419,973.85 & 339,317.90 & \textbf{308,004.50} & 309,195.95  \\
37 & 337,913.00 & 464,681.45 & 372,082.80 & \textbf{359,501.40} & 359,770.00  \\
38 & 319,905.00 & 443,159.95 & 343,042.05 & \textbf{342,613.85} & 342,809.45  \\
39 & 366,848.00 & 534,803.25 & 418,399.85 & 397,048.95 & \textbf{393,949.00}  \\
40 & 305,011.00 & 498,094.70 & 328,389.55 & 321,507.00 & \textbf{317,653.55}  \\ \midrule
median & 245,536.00 & 373,240.90 & 280,093.88 & 261,350.27 & 261,970.30  \\
mean & 263,157.33 & 419,019.72 & 310,837.55 & 285,878.95 & 284,825.75  \\
stand. dev. & 94,057.58 & 194,238.04 & 133,327.52 & 105,938.42 & 105,897.62  \\ [0.5em]
median $\cal G$ &  & 45.17 & 11.70 & 8.14 & 7.09   \\
mean $\cal G$ &  & 55.38 & 16.55 & 8.18 & 7.74   \\
stand. dev.  $\cal G$ &  & 26.49 & 14.65 & 3.02 & 2.92   \\ [0.5em]
no. of best known  & & 0 & 8 & 12 & 20   \\ \bottomrule
\end{tabular}

\end{table}

\begin{table}
\centering
\caption{Results per instance of class $l$ with two agents ($|A|=2$)}
\label{tab:res_l2}
\vspace{5pt}

\footnotesize
\begin{tabular}{rrrrrr} \toprule
 & \multicolumn{1}{c}{$|A| = 1$} & \multicolumn{4}{c}{distributed to $|A| = 2$ agents} \\ \cmidrule(lr){2-2} \cmidrule(lr){3-6}
\multicolumn{1}{l}{class $l$ instance} & \multicolumn{1}{c}{best-known} & \multicolumn{1}{c}{ES} & \multicolumn{1}{c}{SA} & \multicolumn{1}{c}{ACO} & \multicolumn{1}{c}{CACM} \\ \midrule
1 & 591,585.00 & 695,337.65 & 608,572.30 & 642,747.80 & \textbf{597,566.10}  \\
2 & 816,043.00 & 923,305.30 & 885,146.00 & \textbf{878,505.40} & 982,418.75  \\
3 & 908,616.00 & 1,113,228.95 & \textbf{975,174.80} & 1,000,429.40 & 996,370.50  \\
4 & 929,929.00 & 1,037,113.85 & 1,035,097.40 & 1,030,539.85 & \textbf{1,011,923.30}  \\
5 & 1,145,749.00 & 1,371,152.00 & 1,255,854.25 & \textbf{1,222,565.20} & 1,266,812.85  \\
6 & 7,639,920.00 & 12,781,176.60 & 9,782,048.90 & 8,510,346.95 & \textbf{8,198,323.90}  \\
7 & 3,921,407.00 & 5,785,793.10 & 5,275,859.90 & 4,327,733.60 & \textbf{4,054,305.25}  \\
8 & 2,694,924.00 & 3,586,973.70 & 3,235,243.15 & 2,870,682.50 & \textbf{2,824,140.80}  \\
9 & 1,880,021.00 & 2,453,164.65 & 2,061,679.05 & 2,128,470.80 & \textbf{1,899,801.85}  \\
10 & 1,502,194.00 & 1,901,132.75 & 1,624,990.25 & 1,691,979.45 & \textbf{1,569,679.65}  \\
11 & 59,121,845.00 & 120,529,283.50 & 73,444,728.85 & 64,597,231.15 & \textbf{59,407,117.90}  \\
12 & 13,422,827.00 & 22,643,066.50 & 15,480,871.65 & 14,510,728.55 & \textbf{13,573,477.20}  \\
13 & 4,718,146.00 & 7,363,067.15 & 5,867,821.10 & 5,246,854.30 & \textbf{5,008,244.80}  \\
14 & 2,908,634.00 & 3,819,216.00 & 3,195,148.95 & 3,202,206.60 & \textbf{3,068,236.60}  \\
15 & 1,737,525.00 & 2,198,085.20 & 1,853,347.30 & 1,962,320.40 & \textbf{1,838,495.35}  \\
16 & 468,463,630.00 & 959,595,328.50 & 555,959,053.05 & 520,424,836.30 & \textbf{473,526,552.45}  \\
17 & 18,677,678.00 & 41,742,064.65 & 21,357,104.50 & 20,234,057.05 & \textbf{18,860,888.70}  \\
18 & 7,308,193.00 & 12,586,168.40 & 8,640,468.25 & 7,917,765.70 & \textbf{7,526,011.90}  \\
19 & 3,519,932.00 & 4,965,055.40 & 4,028,118.60 & 3,947,226.30 & \textbf{3,713,975.95}  \\
20 & 2,278,214.00 & 2,803,612.00 & 2,446,637.20 & 2,549,066.10 & \textbf{2,401,407.05}  \\
21 & 1,187,090.00 & 1,407,774.55 & 1,323,941.05 & 1,303,687.95 & \textbf{1,243,593.70}  \\
22 & 1,341,584.00 & 1,559,006.70 & \textbf{1,474,146.65} & 1,493,191.95 & 1,536,134.45  \\
23 & 1,400,480.00 & 1,687,215.15 & 1,518,462.70 & 1,548,463.70 & \textbf{1,502,495.80}  \\
24 & 1,382,150.00 & 1,639,929.55 & 1,533,006.85 & 1,563,371.50 & \textbf{1,499,628.70}  \\
25 & 1,657,248.00 & 1,986,023.70 & 1,939,231.90 & 1,859,094.20 & \textbf{1,814,371.90}  \\
26 & 12,671,808.00 & 25,125,549.90 & 15,998,629.30 & 14,568,749.65 & \textbf{13,270,503.10}  \\
27 & 7,159,416.00 & 12,131,464.00 & 8,501,797.65 & 8,461,161.55 & \textbf{7,496,052.05}  \\
28 & 4,148,783.00 & 5,792,328.05 & 4,824,012.55 & 4,770,983.15 & \textbf{4,423,193.05}  \\
29 & 2,889,151.00 & 3,776,572.00 & 3,117,732.50 & 3,339,847.50 & \textbf{3,020,055.65}  \\
30 & 2,183,815.00 & 2,732,881.40 & 2,378,237.60 & 2,588,377.55 & \textbf{2,315,291.70}  \\
31 & 101,497,679.00 & 278,824,944.75 & 128,549,165.95 & 127,935,080.15 & \textbf{102,255,225.00}  \\
32 & 18,028,225.00 & 37,609,906.05 & 20,189,757.75 & 21,578,019.85 & \textbf{18,287,579.60}  \\
33 & 6,780,986.00 & 11,353,745.60 & 10,902,146.20 & 8,401,813.80 & \textbf{7,318,975.85}  \\
34 & 4,055,536.00 & 5,341,405.15 & 5,029,223.90 & 4,927,903.20 & \textbf{4,283,010.35}  \\
35 & 2,559,885.00 & 3,307,932.90 & 2,810,070.40 & 3,050,680.85 & \textbf{2,771,436.85}  \\
36 & 755,506,278.00 & 2,036,463,721.10 & 955,384,033.75 & 863,134,800.00 & \textbf{768,264,766.55}  \\
37 & 33,309,777.00 & 86,653,413.25 & 45,203,262.65 & 37,584,305.00 & \textbf{33,724,916.75}  \\
38 & 10,464,662.00 & 19,257,108.00 & 12,772,705.35 & 11,915,138.85 & \textbf{10,734,284.00}  \\
39 & 5,116,338.00 & 7,422,470.10 & 6,072,238.50 & 6,041,357.80 & \textbf{5,441,242.45}  \\
40 & 3,391,440.00 & 4,574,457.65 & 3,666,782.10 & 4,133,474.95 & \textbf{3,515,181.75}  \\ \midrule
median & 3,455,686.00 & 4,769,756.53 & 3,847,450.35 & 4,040,350.63 & 3,614,578.85  \\
mean & 39,522,983.58 & 93,963,529.39 & 48,805,038.77 & 44,977,394.91 & 40,176,092.25  \\
stand. dev. & 136,399,204.84 & 347,139,859.03 & 169,755,185.52 & 154,825,527.76 & 138,439,704.88  \\ [0.5em]
median $\cal G$ &  & 33.99 & 14.39 & 12.16 & 5.10   \\
mean $\cal G$ &  & 55.07 & 16.84 & 13.30 & 5.43   \\
stand. dev.  $\cal G$ &  & 43.75 & 10.62 & 4.72 & 3.96   \\ [0.5em]
no. of best known  & & 0 & 2 & 2 & 36   \\ \bottomrule
\end{tabular}
\end{table}

\begin{table}
\centering
\caption{Results per instance of class $l$ with five agents ($|A|=5$)}
\label{tab:res_l5}
\vspace{5pt}
\footnotesize
\begin{tabular}{rrrrrr} \toprule
 & \multicolumn{1}{c}{$|A| = 1$} & \multicolumn{4}{c}{distributed to $|A| = 5$ agents} \\ \cmidrule(lr){2-2} \cmidrule(lr){3-6}
\multicolumn{1}{l}{class $l$ instance} & \multicolumn{1}{c}{best-known} & \multicolumn{1}{c}{ES} & \multicolumn{1}{c}{SA} & \multicolumn{1}{c}{ACO} & \multicolumn{1}{c}{CACM} \\ \midrule
1 & 591,585.00 & 804,893.65 & 710,850.55 & 647,247.70 & \textbf{635,739.60}  \\
2 & 816,043.00 & 1,147,738.65 & 988,143.20 & 930,055.95 & \textbf{821,528.70}  \\
3 & 908,616.00 & 1,355,186.90 & 1,188,264.05 & 1,084,152.75 & \textbf{948,842.40}  \\
4 & 929,929.00 & 1,247,533.15 & 1,184,439.70 & 1,128,669.65 & \textbf{967,533.50}  \\
5 & 1,145,749.00 & 1,413,376.55 & 1,259,973.25 & 1,399,061.45 & \textbf{1,205,951.95}  \\
6 & 7,639,920.00 & 25,878,434.65 & 19,037,568.40 & 8,694,605.45 & \textbf{8,355,460.70}  \\
7 & 3,921,407.00 & 9,132,315.80 & 8,707,302.60 & 4,653,392.50 & \textbf{4,237,945.50}  \\
8 & 2,694,924.00 & 4,358,121.20 & 3,438,619.75 & 3,346,616.65 & \textbf{2,947,906.85}  \\
9 & 1,880,021.00 & 2,710,619.00 & 2,137,343.75 & 2,336,763.45 & \textbf{2,028,723.05}  \\
10 & 1,502,194.00 & 2,085,083.65 & 1,677,771.85 & 1,867,070.70 & \textbf{1,583,783.40}  \\
11 & 59,121,845.00 & 137,258,807.55 & 98,380,736.30 & 74,420,057.30 & \textbf{59,486,366.90}  \\
12 & 13,422,827.00 & 31,575,365.70 & 18,208,444.20 & 17,050,429.70 & \textbf{13,619,454.45}  \\
13 & 4,718,146.00 & 8,582,950.70 & 6,387,892.00 & 5,742,169.60 & \textbf{5,117,738.80}  \\
14 & 2,908,634.00 & 4,513,204.40 & 3,650,417.90 & 3,651,103.30 & \textbf{3,172,980.55}  \\
15 & 1,737,525.00 & 2,549,510.10 & 1,951,706.95 & 2,266,967.25 & \textbf{1,828,004.20}  \\
16 & 468,463,630.00 & 1,173,231,850.40 & 724,511,376.10 & 556,757,116.65 & \textbf{473,495,284.60}  \\
17 & 18,677,678.00 & 48,122,890.00 & 31,079,449.25 & 20,942,784.70 & \textbf{18,866,385.50}  \\
18 & 7,308,193.00 & 15,131,195.70 & 10,952,393.35 & 8,489,683.80 & \textbf{7,641,505.35}  \\
19 & 3,519,932.00 & 5,773,626.55 & 4,671,451.50 & 4,434,413.15 & \textbf{3,811,046.35}  \\
20 & 2,278,214.00 & 3,433,568.70 & 2,855,260.20 & 3,021,621.50 & \textbf{2,466,302.90}  \\
21 & 1,187,090.00 & 1,789,713.70 & 1,618,514.00 & 1,393,015.70 & \textbf{1,232,991.95}  \\
22 & 1,341,584.00 & 1,970,678.35 & 1,660,601.70 & 1,637,945.90 & \textbf{1,400,411.50}  \\
23 & 1,400,480.00 & 1,902,855.80 & 1,694,771.35 & 1,666,318.55 & \textbf{1,416,472.45}  \\
24 & 1,382,150.00 & 1,889,928.60 & 1,583,299.20 & 1,780,922.50 & \textbf{1,406,540.30}  \\
25 & 1,657,248.00 & 2,103,313.15 & 1,987,959.25 & 2,070,126.30 & \textbf{1,739,024.90}  \\
26 & 12,671,808.00 & 34,326,086.25 & 25,969,790.05 & 16,071,795.85 & \textbf{13,304,507.20}  \\
27 & 7,159,416.00 & 15,489,176.70 & 12,806,323.95 & 9,074,157.40 & \textbf{7,780,993.10}  \\
28 & 4,148,783.00 & 7,384,138.30 & 5,620,887.95 & 5,449,223.35 & \textbf{4,657,224.70}  \\
29 & 2,889,151.00 & 4,480,227.95 & 3,571,646.00 & 3,882,516.70 & \textbf{3,011,682.40}  \\
30 & 2,183,815.00 & 3,160,428.90 & 2,567,987.70 & 2,859,877.00 & \textbf{2,264,261.75}  \\
31 & 101,497,679.00 & 417,268,369.80 & 302,711,993.50 & 128,419,919.00 & \textbf{102,344,312.65}  \\
32 & 18,028,225.00 & 58,285,339.85 & 35,763,440.10 & 23,474,329.35 & \textbf{18,573,788.45}  \\
33 & 6,780,986.00 & 14,718,304.85 & 11,170,886.30 & 8,494,852.00 & \textbf{7,396,134.75}  \\
34 & 4,055,536.00 & 7,075,030.85 & 4,873,896.90 & 5,375,282.75 & \textbf{4,344,803.30}  \\
35 & 2,559,885.00 & 4,043,062.15 & 2,968,750.20 & 3,475,333.80 & \textbf{2,680,782.55}  \\
36 & 755,506,278.00 & 2,594,443,728.80 & 1,259,991,223.40 & 863,673,114.45 & \textbf{768,319,555.25}  \\
37 & 33,309,777.00 & 101,154,724.65 & 51,080,825.70 & 37,632,659.60 & \textbf{33,725,180.95}  \\
38 & 10,464,662.00 & 24,143,971.55 & 15,436,884.20 & 12,386,857.75 & \textbf{10,804,778.05}  \\
39 & 5,116,338.00 & 9,116,234.95 & 8,180,743.30 & 6,700,266.75 & \textbf{5,678,066.85}  \\
40 & 3,391,440.00 & 5,232,692.40 & 3,935,280.35 & 4,556,441.20 & \textbf{3,557,098.90}  \\ \midrule
median & 3,455,686.00 & 5,503,159.48 & 4,303,365.93 & 4,495,427.18 & 3,684,072.63  \\
mean & 39,522,983.58 & 119,757,107.01 & 67,454,377.75 & 46,573,473.48 & 40,221,927.43  \\
stand. dev. & 136,399,204.84 & 440,099,405.84 & 225,778,836.85 & 157,709,294.72 & 138,436,787.71  \\ [0.5em]
median $\cal G$ &  & 62.87 & 31.75 & 24.60 & 4.80   \\
mean $\cal G$ &  & 95.93 & 46.38 & 23.79 & 5.17   \\
stand. dev.  $\cal G$ &  & 69.34 & 39.91 & 6.76 & 3.15   \\ [0.5em]
no. of best known  & & 0 & 0 & 0 & 40   \\ \bottomrule
\end{tabular}
\end{table}


\end{document}